\documentclass[a4paper]{llncs}

\usepackage{amssymb}
\usepackage{graphicx}
\usepackage{epstopdf}
\usepackage{color}
\usepackage{bm}
\usepackage{upgreek}
\usepackage{hyperref}

\usepackage{url}
\urldef{\mailsa}\path|{amirj,az}@robots.ox.ac.uk|
\urldef{\mailsb}\path|timor.kadir@optellum.com|

\def\presec{\vspace{-5pt}}
\def\presubsec{\vspace{-5pt}}
\def\postsec{\vspace{-5pt}}

\usepackage{bm}
\usepackage{amsmath}

\usepackage{algorithm}
\usepackage{program}
\usepackage{multirow}
\usepackage{booktabs}
\begin{document}
	
	\mainmatter
	\title{Self-Supervised Learning for Spinal MRIs}
	
	\author{Amir Jamaludin\inst{1}\and Timor Kadir\inst{2}\and Andrew Zisserman\inst{1}}
	\institute{VGG, Dept.\ of Engineering Science, University of Oxford
		\and
		Optellum\\
		\mailsa\\
		\mailsb\\}
	\maketitle
	
	\begin{abstract}
A significant proportion of patients scanned in a clinical setting
have follow-up scans. We show in this work that such longitudinal
scans alone can be used as a form of ``free'' self-supervision for
training a deep network. We demonstrate this self-supervised learning
for the case of T2-weighted sagittal lumbar Magnetic Resonance Images
(MRIs). A Siamese convolutional neural network (CNN) is trained using
two losses: (i) a contrastive loss on whether the scan is of the same
person (i.e.\ longitudinal) or not, together with (ii) a
classification loss on predicting the level of vertebral bodies. The
performance of this pre-trained network is then assessed on a grading
classification task. We experiment on a dataset of 1016 subjects, 423
possessing follow-up scans, with the end goal of learning the disc
degeneration radiological gradings attached to the intervertebral
discs. We show that the performance of the pre-trained CNN on the
supervised classification task is (i) superior to 
that of a network trained from scratch;  and (ii) requires far
fewer annotated training samples to reach an equivalent performance to
that of the network trained from scratch.  
\end{abstract}
	
	\presec \section{Introduction}
	A prerequisite for the utilization of machine learning methods in medical image understanding problems is the collection of suitably curated and annotated clinical datasets for training and testing. Due to the expense of collecting large medical datasets with the associated ground-truth, it is important to develop new techniques to maximise the use of available data and minimize the effort required to collect new cases.  
	
In this paper, we propose a self-supervision approach that can be used
to pre-train a CNN using the embedded information that is readily
available with standard clinical image data. 
Many patients are scanned multiple times in a so-called
longitudinal manner, for instance to assess changes in disease state
or to monitor therapy. We define a pre-training scheme using only the
information about which scans belong to the same patient. Note, 
we do not need to know the identity of the patient; only which images
belong to the same patient. This information is readily available 
in formats such as DICOM
(Digital Imaging and Communications in Medicine) that typically include a
rich set of meta-data such as patient identity, date of birth and
imaging protocol (and 
DICOM anonymization software typically
assigns the same `fake-id' to images of the same patient).

Here,  we implement this  self-supervision pre-training for the case of 
T2-weighted sagittal lumbar MRIs. 
We train a Siamese CNN to distinguish between pairs of images that
contain the same patient scanned at different points in time,  and
pairs of images of entirely different patients.
We also illustrate that additional data-dependent
self-supervision tasks can be included by specifying  
an auxiliary task of predicting vertebral body levels, and including both types of self-supervision
in a multi-task training scheme.
Following the pre-training, the learned weights are transferred to a
classification CNN that, in our particular application of interest, is
trained to predict a range of gradings related to vertebra and disc
degeneration.  The performance of the classification CNN is then used to evaluate
the success of the pre-training.  We also compare 
to  pre-training
a CNN on a large dataset of lumbar T2-weighted
sagittal MRIs  fully annotated with eight radiological gradings~\cite{Jamaludin16Short}.
	
	\vspace*{1.5mm} \noindent \textbf{Related Work:}
Models trained using only information contained within an image as a
supervisory signal have been proven to be effective feature
descriptors e.g.\ \cite{Doersch15} showed that a CNN trained to
predict relative location of pairs of patches, essentially learning
spatial context, is better at a classification task (after
fine-tuning) then a CNN trained from scratch.  
The task of learning scans from
the same unique identity is related to slow-feature learning in
videos~\cite{mobahi2009deep,wang2015unsupervised,wiskott2002slow}. 

Transfer
learning in CNNs has been found to be extremely effective especially
when the model has been pre-trained on a large dataset like ImageNet~\cite{Deng09}, and there have
been several successes on using models pre-trained on natural images
on medical images e.g.\  lung disease classification \cite{Shin16}. 
However, since a substantial portion of medical images
are volumetric in nature,  it is also appropriate to
experiment with transfer learning with scans of the same modality
rather than using ImageNet-trained models. 
	
	\presec\section{Self-Supervised Learning}
	
	This section first describes the details of the input volumes
	extraction:  the vertebral bodies ({\bf VBs}) that will be used for the
	self-supervised training; and  the intervertebral discs ({\bf IVDs}) that
	will be used for the supervised classification
	experiments. We then describe the loss functions and network architecture.
	
	\presubsec\subsection{Extracting Vertebral Bodies and Intervertebral Discs} \label{input}
\begin{figure}[h!]
		\centering
		\includegraphics[width=1.0\textwidth]{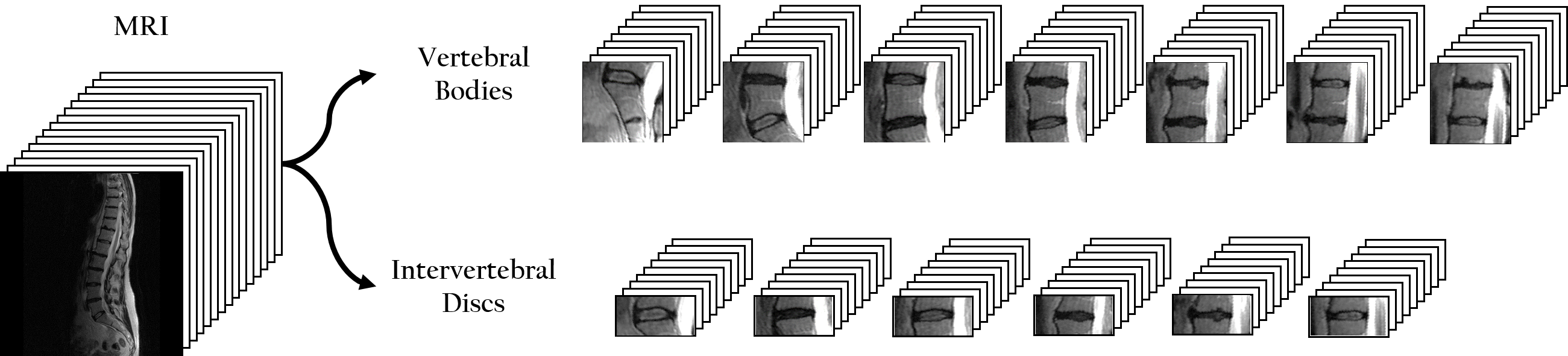}
		\caption{\textbf{Extracting VB and IVD Volumes:} For each MRI, we extract 7 VB (T12 to S1) and 6 IVD (T12-L1 to L5-S1) volumes. The dimensions of the volumes are: (i) $224\times224\times9$ for the VBs, and (ii) $112\times224\times9$ for the IVDs. The whole volume is centred at the detected middle slice of the volume of the VB/IVD.}
		\label{fig:inputs}
	\end{figure}

	For each T2-weighted sagittal MRI we automatically detect bounding volumes of the (T12 to S1) VBs alongside the level labels using the pipeline outlined in \cite{Jamaludin16Short,Lootus13Short}. As per \cite{Jamaludin16Short}, we also extract the corresponding IVD volumes (T12-L1 to L5-S1, where T, L, and S refer to the thoracic, lumbar, and sacral vertebrae) from the pairs of vertebrae e.g. a L5-S1 IVD is the disc between the L5 and S1 vertebrae. Fig.~\ref{fig:inputs} shows the input and outputs of the extraction pipeline. The slices of both the VB and IVD volumes are mid-sagittally aligned (to prevent misalignment that can occur from scoliosis and other disorders) and zero-padded slice-wise if the number of slices is below the predefined 9 channels. The volumes are rescaled according to the width of the VB or IVD and normalized such that the median intensity of the VB above and below the current VB or IVD is $0.5$.
	
	\presubsec\subsection{Longitudinal Learning via Contrastive Loss} \label{llearning}
	
\begin{figure}[h!]
		\centering
		\includegraphics[width=1.0\textwidth]{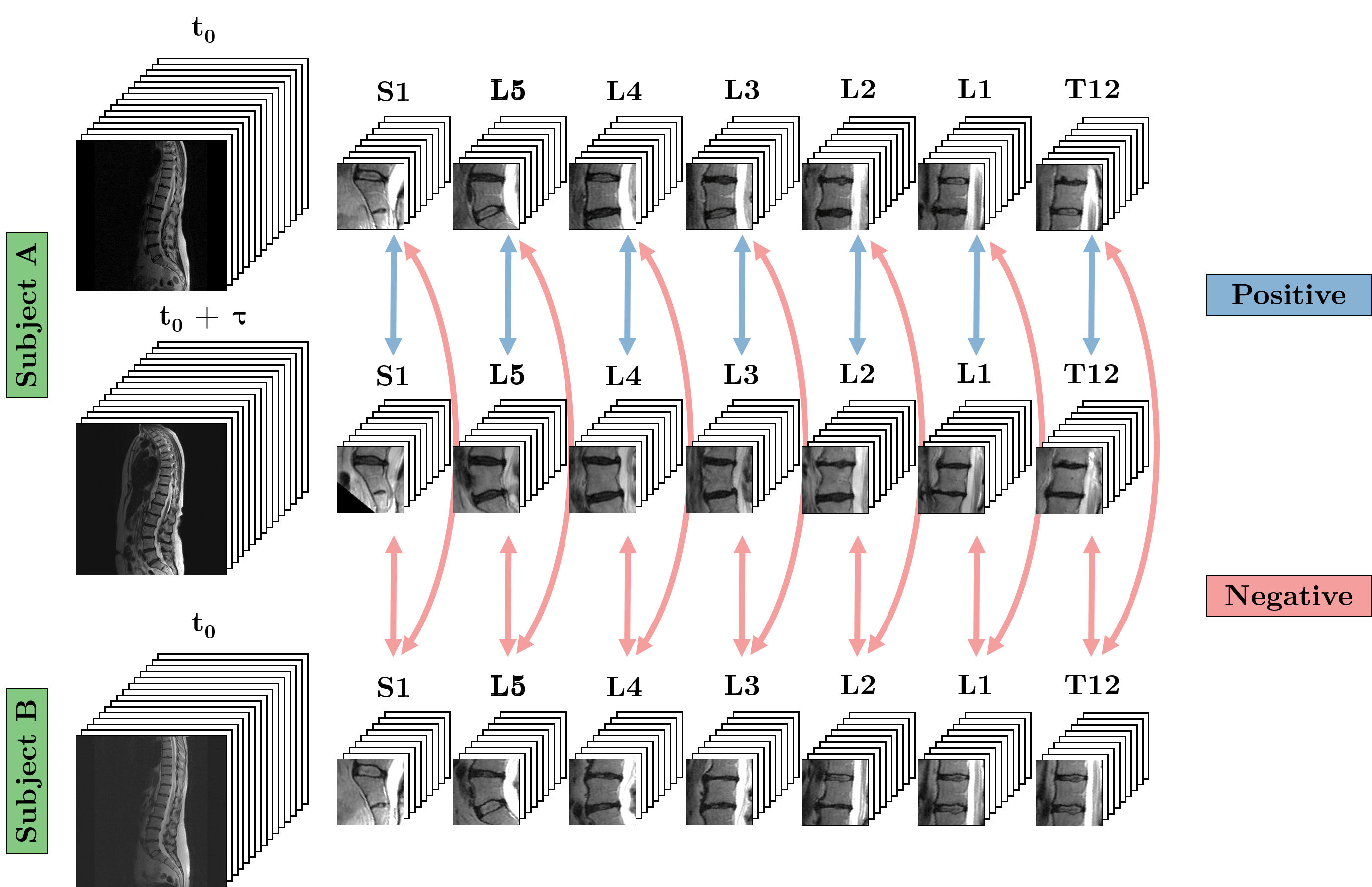}
		\caption{\textbf{Generating the positive/negative pairs}: The arrows mark a pair of VBs, where blue arrows highlight positive pairs while negative are highlighted in red. Each pair is generated from two scans. $\mathbf{t_0}$ refers to the time of the initial scan and $\boldsymbol{\uptau}$  is the time between the baseline and the follow-up scans, typically 10 years in our dataset.}
		\label{fig:sampling}
	\end{figure}

	The longitudinal information of
	the scans is used to train a Siamese network such that the embeddings for scans of the same person are close, whereas scans
	of different people are not. The input is a pair of VBs of
	the same level; an S1 VB is only compared against an S1 VB and vice
	versa. We use the contrastive loss in \cite{Chopra05}, $\mathcal{L}_C = \sum_{n=1}^{N} (y)d^2 + (1-y)\max(0,m - d)^2$, 
	where $d = \| a_n - b_n \|_2$, and  $a_n$ and $b_n$ are the 1024-dimensional \textbf{FC7} (embedding) vectors for the first and second VB in an input pair, and $m$ is a predefined margin.
	Positive, $y = 1$, VB pairs are those that were obtained
	from a single unique subject (same VB scanned at different points in
	time) and negative, $y = 0$, pairs are VBs from different individuals
	(see Fig.~\ref{fig:sampling}). 
	
\begin{figure}[h!]
		\centering
		\includegraphics[width=1.0\textwidth]{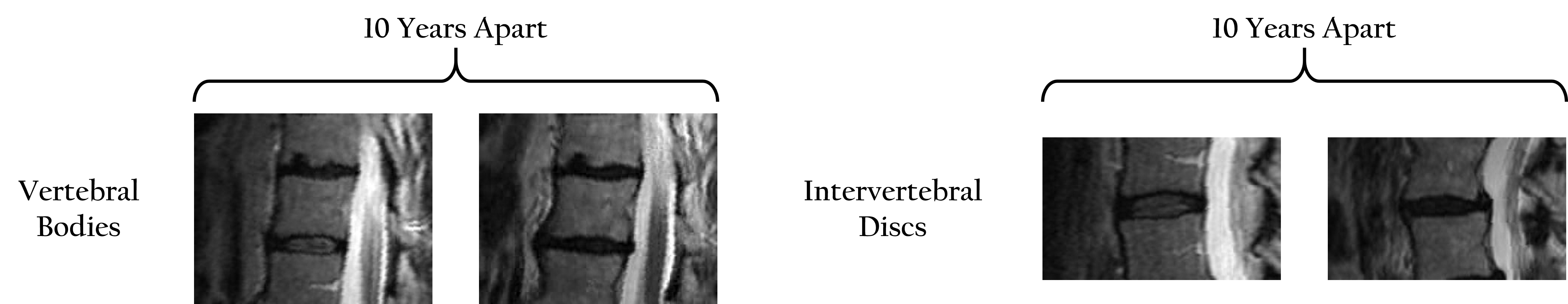}
		\caption{\textbf{VB and IVD Across Time}: The VB appears unchanged but over time the IVD loses intensity of its nucleus and experiences a loss in height. Furthermore, in the IVD example, we can observe a vertebral slip, or spondylolisthesis, which does not change the appearance of the VBs themselves but significantly changes the IVD.}
		\label{fig:whyVB}
		\vspace{-4mm} 
	\end{figure}

	We use the VBs instead of the IVDs due to the fact that vertebrae tend to be more constant in shape and appearance over time. Fig.~\ref{fig:whyVB} shows examples of both VB and IVD at different points in time. In other medical tasks the pair of VBs can easily be changed to other anatomies for example comparing lungs in chest X-rays.
	
	\vspace{-6mm} 
	
	\presubsec\subsection{Auxiliary Loss -- Predicting VB Levels} \label{prox}	
	In addition to the contrastive loss, we also employ an auxiliary 	loss to give complementary supervision.
	Since each
	VB pair is made up of VBs of the same level, we train a classifier on
	top of the \textbf{FC7} layer, i.e.\ the discriminative layer, to
	predict the seven levels of the VB (from T12 to S1). The overall loss
	can then be described as a combination of the contrastive and softmax
	log losses, {$\mathcal{L} = \mathcal{L}_C
		-\sum_{n=1}^{N}\alpha_c\left(y_c(x_n) - \log\sum_{j=1}^{7}
		e^{y_j(x_n)}\right)$}, where $y_j$ is the $j$th component of the
	\textbf{FC8} output, $c$ is the true class of $x_n$, and $\alpha_c$ is
	the class-balanced weight as in \cite{Jamaludin16Short}.
	
	\presubsec\subsection{Architecture} \label{classification}	
\begin{figure}[h!]
		\centering
		\includegraphics[width=1.0\textwidth]{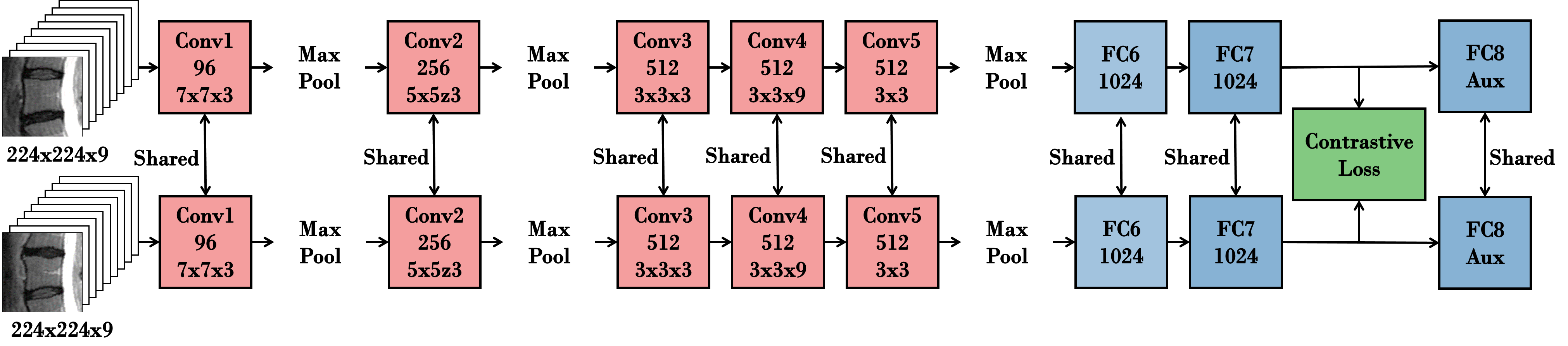}
		\caption{The Siamese network architecture is trained to distinguish between VB pairs from the same subject (same VB scanned a decade apart) or VB pairs from different subjects. There is also an additional loss (shown as aux) to predict the level of the VB.}
		\label{fig:arch}
	\end{figure}

	The base architecture trained to distinguish VB pairs is based of the VGG-M network in \cite{Chatfield14} (see Fig.~\ref{fig:arch}). The input to the Siamese CNN is the pairs of VBs, with dimension $224 \times 224 \times 9$. We use 3D kernels from \textbf{Conv1} to \textbf{Conv4} layers followed by a 2D \textbf{Conv5}. To transform the tensor to be compatible with 2D kernels, the \textbf{Conv4} kernel is set to be $3 \times 3 \times 9$ with no padding, resulting in a reduction of the slice-wise dimension after \textbf{Conv4}. We use $2\times2$ max pooling. The output dimension of the \textbf{FC8} layer depends on the number of classes i.e. seven for the self-supervisory auxiliary task of predicting the seven VB levels.
	
	\presec\section{Dataset \& Implementation Details} \label{sec:implementation}	
	
	We experimented on a dataset of 1016 recruited subjects (predominantly
	female) from the TwinsUK registry (\url{www.twinsuk.ac.uk}) not using
	backpain as an exclusion or selection criterion. As well as a
	baseline scan for each subject, 423 subjects have follow-up scans
	taken 8-12 years after the original baseline.
	A majority of the
	subjects with the follow-up scans have only two scans (one baseline
	and one follow-up) while a minority have three (one baseline and two
	follow-up). 
	The baseline scans were
	taken with a 1.0-Tesla scanner while the follow-up scans were taken
	with a 1.5-Tesla machine but both adhered to the same scanning
	protocol (slice thickness, times to recovery and echo, TR and
	TE). Only T2-weighted sagittal scans were collected for each subject.

	\presubsec\subsection{Radiological Gradings} \label{gradings}
	
\begin{figure}[h!]
		\centering
		\includegraphics[width=1.0\textwidth]{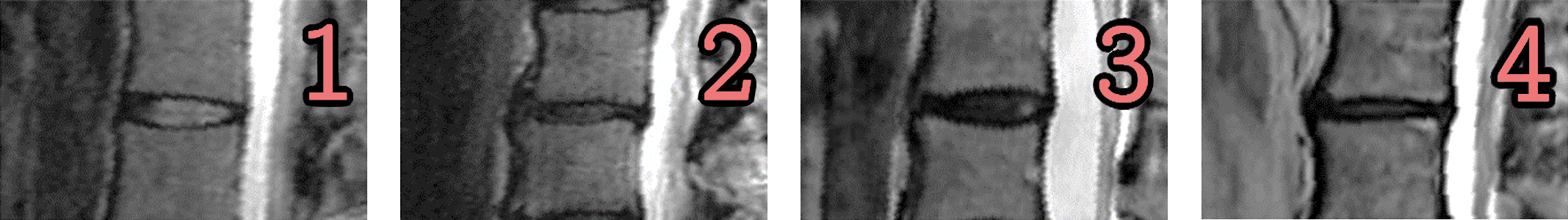}
		\caption{\textbf{Disc Degeneration}: A four-class grading system based on Pfirrmann grading that depends on the intensity and height of the disc.}
		\label{fig:gradings}
	\end{figure}

	The subjects were graded with a measure of
	\textbf{Disc Degeneration}, not dissimilar to Pfirrmann Grading. The
	gradings were annotated by a clinician and were done on a per disc
	basis: from L1-L2 to L5-S1 discs (5 discs per subject). We use these
	gradings to assess the benefits of pre-training a classification
	network on longitudinal information. Examples of the gradings can be
	seen in Fig.~\ref{fig:gradings}.  920 of the 1016  subjects are graded.
	
	\presubsec\subsection{Training} \label{training}
	
	\vspace*{1.5mm} \noindent \textbf{Data augmentation:}
	The augmentation strategies are identical to  that described in \cite{Jamaludin16Short} for the classification CNN while  we use slightly different augmentations to train the Siamese network. The augmentations are:  (i) rotation with $\theta$ = $-15^{\circ}$ to $15^{\circ}$, (ii) translation of $\pm 48$ pixels in the x-axis, $\pm 24$ pixels in the y-axis, $\pm 2$ slices in the z-axis, (iii) rescaling with a scaling factor between $90\%$ to $110\%$, (iv) intensity variation $\pm0.2$, and (v) random slice-wise flip i.e. reflection of the slices across the mid-sagittal (done on a per VB pair basis).
	
	\vspace*{1.5mm} \noindent \textbf{Details:}
	Our implementation is based on the MatConvNet toolbox~\cite{Vedaldi14a} and the networks were trained using an NVIDIA Titan X GPU. The hyperparameters are: batch size 128 for classification and 32 for the Siamese network; momentum 0.9; weight decay 0.0005; learning rate $1e^{-3}$ (classification) and $1e^{-5}$ (self-supervision) and lowered by a factor of 10 as the validation error plateaus, which is also our stopping criterion, normally around 2000 epochs for the classification network and 500 epochs for the Siamese network.

	\presec\section{Experiments \& Results}\label{exp}
	
	\presubsec\subsection{Longitudinal Learning \& Self-supervision} \label{expLL}			
	
\begin{figure}[t!]
		\centering
		\includegraphics[width=0.8\textwidth]{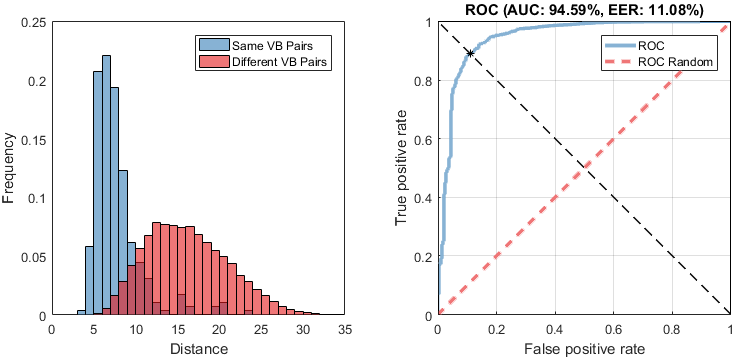}
		\caption{\textbf{Left:} Histogram of VB pairs distances in the test set. Positive pairs in blue, negative in red. \textbf{Right:} The ROC of the classification of positive/negative VB pairs.}
		\label{fig:distances}
	\end{figure}

	The dataset is split by subject 
	80:10:10 into train, validation, and test sets. For the 423 subjects with multiple scans this results in
	a 339:42:42 split. Note, a pair of twins will only be in one set i.e. one subject part of a twin pair can't be in training and the other in test.	With the trained network, each input VB can be represented as a
	1024-dimensional \textbf{FC7} vector. For each pair of VBs i.e.\ two
	\textbf{FC7} vectors, we can calculate the $L_2$ distance between two
	samples, which is the same distance function used during
	training. 
	Fig.~\ref{fig:distances} shows the histogram of the distances (both positive and negative)
	for all the VB pairs in the test set.
	In general, VB pairs that are from the same
	subject have lower distances compared to pairs from different
	subjects. Using the distances between pairs as classification scores
	of predicting whether the VB pairs are from the same or different
	subjects, we achieve an AUC of $\mathbf{94.6}$\textbf{\%}. We also
	obtain a very good performance of $\mathbf{97.8}$\textbf{\%} accuracy
	on the auxiliary task of predicting VB level.
	
	\presubsec\subsection{Benefits of Pre-training} \label{expC}
	
\begin{figure}[h!]
		\centering
		\includegraphics[width=1.0\textwidth]{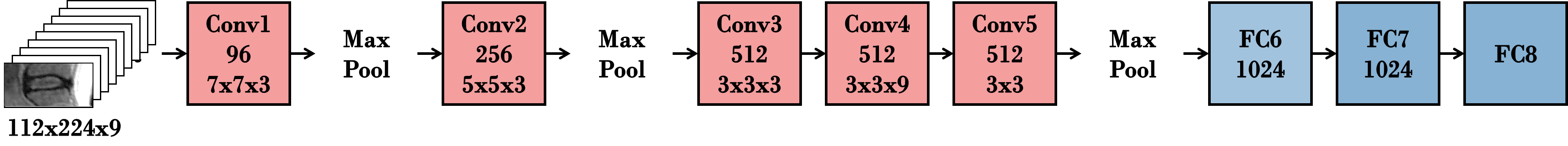}
		\caption{The architecture of the  classification CNN. The convolutional weights are obtained from the Siamese network.}
		\label{fig:class}
	\end{figure}

	To measure the performance gained by pre-training using the
	longitudinal information we 
	use the 
	convolution weights learnt in the Siamese network, and
	train a classification CNN to predict the
	\textbf{Disc Degeneration} radiological grading  (see Fig.~\ref{fig:class}). For this
	classification task we use the 920 subjects that possess gradings and
	split them into the following sets: 670 for training, 50 in validation, and 200 for
	testing. Subjects with follow-up scans ($>1$ scans) are only used in
	training and not for testing so,  in essence, the Siamese network
	will never have seen the subjects in the classification test set.
	
	We transfer and freeze convolutional weights of the Siamese network
	and only train the randomly-initialized fully-connected layers. We
	also experimented with fine-tuning the convolutional layers but we
	find the difference in performance to be negligible. For comparison, we
	also train: (i) a CNN from scratch, (ii) a CNN with a frozen randomly
	initialized convolutional layers (to see the power of just training on
	the fully-connected layers) as a  baseline, and (iii) a CNN using
	convolutional weights of a CNN trained on a fully-annotated spinal MRI
	dataset with multiple radiological gradings in \cite{Jamaludin16Short}.
	The performance measure is the  average class accuracy, calculated as the average of the
	diagonal elements of the normalized confusion matrix.
	
\begin{figure}[h!]
		\centering
		\includegraphics[width=0.8\textwidth]{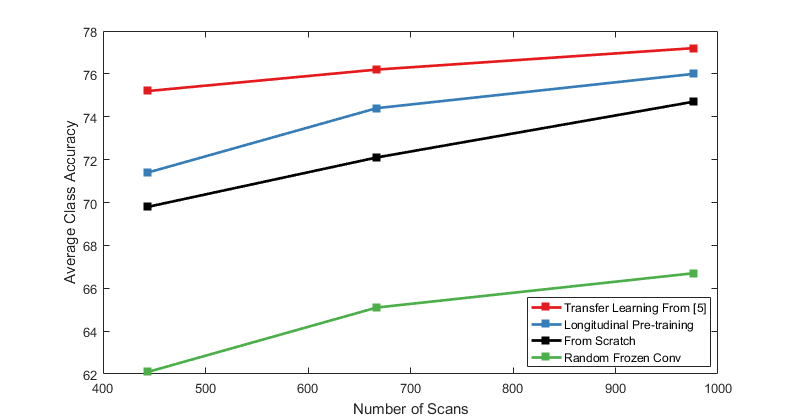}
		\caption{
			Accuracy as the number of training samples for IVD classification is increased.
			Longitudinal pre-training improves over
			training from scratch, showing its benefit, and this performance boost persists even at 976 scans.
			Transfer learning from a CNN trained on
			a strongly-supervised dataset (including IVD classification) is better and provides an `upper bound' 
			on transfer performance. Note, even with
			976 scans in the training set, the performance has not plateaued
			hinting at further improvements  with the availability of  more data.
		}
		\label{fig:results}
	\end{figure}

	Fig.~\ref{fig:results} shows the performance of all the models as the
	number of training samples is varied at [240, 361, 670] subjects or
	[444, 667, 976] scans.  It can be seen that with longitudinal
	pre-training, less data is required to reach an equivalent point to
	that of training from scratch, e.g.\ the performance is
	$\mathbf{74.4}$\textbf{\%} using only 667 scans by pre-training,
	whereas training from scratch requires 976 scans to get to
	$\mathbf{74.7}$\textbf{\%}. This performance gain can also be seen
	with a lower amount of training data. As would be expected, transfer
	learning from a CNN trained with strong supervision using the data in \cite{Jamaludin16Short} is better,
	with an accuracy  at least $\mathbf{2.5}$\textbf{\%} above training from
	scratch. Unsurprisingly, a classifier
	trained on top of fully random convolutional weights performs the
	worst. 
	
	Since longitudinal information is essentially
	freely available when collecting data, the performance gain from
	longitudinal pre-training is also free. 
	It is interesting to note that even though the Siamese network
	is trained on a slightly different input, VBs instead of IVDs,
	transferring the weights for classification task on IVDs still achieves better
	performance than starting from scratch. However, we suspect this
	difference in inputs is the primary reason why the performance gain is
	lower when the amount of training data is low as the network needs
	more data to adapt to the IVD input/task (in contrast, the strongly supervised CNN is trained on both VB and IVD classification).

	\presec\section{Conclusion}\postsec
	We have shown that it is possible to use self-supervision to improve performance on a radiological grading classification task and we hope to explore the benefits of adding more auxiliary tasks in the near future e.g. predicting gender, age and weight. The performance improvement is nearing that of transfer learning from a model trained on a fully annotated dataset given that the target training set itself contains enough data. Furthermore, having  a distance function between vertebral pairs opens up the possibility of identifying people using their MRIs.
	
	\vspace{-5pt}
	{\small \subsubsection*{Acknowledgements.}  We are grateful for discussions with Prof.\ Jeremy Fairbank, Dr.\ Jill Urban, and Dr.\ Frances Williams. This work was supported by the RCUK CDT in Healthcare Innovation (EP/G036861/1) and the EPSRC Programme Grant Seebibyte (EP/M013774/1). TwinsUK is funded by the Wellcome Trust; European Community’s Seventh Framework Programme (FP7/2007-2013). The study also receives support from the National Institute for Health Research (NIHR) Clinical Research Facility at Guy’s \& St Thomas’ NHS Foundation Trust and NIHR Biomedical Research Centre based at Guy's and St Thomas' NHS Foundation Trust and King's College London.}
	
	\bibliographystyle{splncs03}
	\bibliography{Bib/shortstrings,Bib/vgg_local,Bib/vgg_other,Bib/mybib}
\end{document}